\documentclass[12pt]{article}

\usepackage{amsthm}
\usepackage[T1]{fontenc}
\usepackage{newtxtext}
\usepackage{newtxmath}
\usepackage{bm}

\usepackage[parfill]{parskip}
\usepackage{afterpage}

\usepackage{setspace}

\RequirePackage[bf,tiny]{titlesec}

\usepackage[usenames,dvipsnames]{xcolor}

\usepackage{graphicx}
\usepackage[labelfont=bf]{caption}
\usepackage[format=hang]{subcaption}

\usepackage{listings}
\usepackage{fancyvrb}
\fvset{fontsize=\normalsize}

\lstdefinestyle{mystyle}{
    commentstyle=\color{OliveGreen},
    keywordstyle=\color{BurntOrange},
    numberstyle=\tiny\color{black!60},
    stringstyle=\color{darkblue},
    basicstyle=\ttfamily,
    breakatwhitespace=false,
    breaklines=true,
    captionpos=b,
    keepspaces=true,
    numbers=left,
    numbersep=5pt,
    showspaces=false,
    showstringspaces=false,
    showtabs=false,
    tabsize=2
}
\usepackage{enumitem}

\usepackage{natbib}
\usepackage[colorlinks,linktoc=all]{hyperref}
\usepackage[all]{hypcap}
\hypersetup{citecolor=MidnightBlue}
\hypersetup{urlcolor=MidnightBlue}
\hypersetup{linkcolor=black}

\usepackage[nameinlink]{cleveref}
\creflabelformat{equation}{#1#2#3}
\crefname{equation}{eq.}{eqs.}
\Crefname{equation}{Eq.}{Eqs.}
\Crefname{section}{\S}{\S}

\usepackage
[acronym,smallcaps,nowarn,section,nogroupskip,nonumberlist]{glossaries}
\glsdisablehyper{}
\setglossarystyle{long3col}
\makeglossaries

\usepackage{blindtext}

\DeclareRobustCommand{\parhead}[1]{\textbf{#1}~}

\newtheoremstyle{simple}
{10pt}{10pt}
{}{}
{\bfseries}{.}
{.5em}
{\thmname{#1}\thmnumber{ #2: }\thmnote{#3}}

\theoremstyle{simple}

\newtheorem{definition}{Definition}
\Crefname{definition}{Definition}{Definitions}

\newtheorem{assumption}{Assumption}
\Crefname{assumption}{Assumption}{Assumptions}

\newtheorem{theorem}{Theorem}
\Crefname{theorem}{Theorem}{Theorems}

\newcommand{\g}{\,\vert\,}

\newcommand{\indpt}{\protect\mathpalette{\protect\independenT}{\perp}}
\def\independenT#1#2{\mathrel{\rlap{$#1#2$}\mkern2mu{#1#2}}}

\newcommand\dif{\mathop{}\!\mathrm{d}}

\newcommand{\mba}{\mathbf{a}}

\newcommand{\mbA}{\mathbf{A}}

\newcommand{\upP}{\mathrm{P}}
\newcommand{\rms}{\mathrm{s}}
\newcommand{\rmm}{\mathrm{*}}

\usepackage{authblk}

\title{\Large Towards Clarifying the Theory of the Deconfounder}
\author[1]{\normalsize Yixin Wang}
\author[2]{\normalsize David M. Blei}

\affil[1]{Department of Statistics; Columbia University}
\affil[2]{Department of Computer Science and Department of Statistics;
  Columbia University}

\date{}

\begin{document}

\maketitle

\begin{abstract}
  \citet{Wang:2019a} studies multiple causal inference and proposes
  the deconfounder algorithm. The paper discusses theoretical
  requirements and presents empirical studies. Several refinements
  have been suggested around the theory of the deconfounder. Among
  these, Imai and Jiang clarified the assumption of ``no unobserved
  single-cause confounders.''  Using their assumption, this paper
  clarifies the theory.  Furthermore, \citet{Ogburn:2020} proposes
  counterexamples to the theory. But the proposed counterexamples do
  not satisfy the required assumptions.
\end{abstract}

\citet{Wang:2019a} studies multiple causal inference and proposes the
deconfounder algorithm. It discusses theoretical requirements and
presents empirical studies. \citet{Wang:2019a} was discussed at JSM
2019 and later in print by \citet{damourcomment},
\citet{imbenscomment}, \citet{imaicomment}, and \citet{ogburncomment};
\citet{Wang:2019b} responds to the comments.

Several refinements have been suggested around the theory of the
deconfounder. Among these, Imai and Jiang clarified the assumption of
``no unobserved single-cause confounders,'' which is discussed in
\citet{Wang:2019b}. Using the refined assumption, this paper discusses
the theory.  Further, in a continuation of their commentary,
\citet{Ogburn:2020} proposes counterexamples to \citet{Wang:2019a}.
This paper also shows how the proposed counterexamples do not satisfy
the required assumptions.

The theoretical results below are the same as those in
\citet{Wang:2019a}.  The original paper constructs the substitute
confounder with a probabilistic factor model and then states the
assumptions under which that construction satisfies weak
unconfoundedness.  This paper first states the assumptions under which
a substitute confounder exists, shows that it satisfies weak
unconfoundedness, and then shows how the deconfounder algorithm
constructs it.

\section{Clarifying the theory of the deconfounder}

Begin by defining a multi-cause separator, a type of random variable.

\begin{definition}[Multi-cause separator]
  Consider all the causes $\mbA = \{A_1, \ldots, A_m\}$.  A
  \textit{multi-cause separator} $U$ is a smallest $\sigma$-algebra
  that renders all the causes conditionally independent,
  \begin{align}
    \label{eq:factor-condindep}
    \upP(A_{1}, \ldots, A_{m}\g U) =
    \prod_{j=1}^m \upP(A_{j}\g U),
  \end{align}
  and where none of the conditionals $\upP(A_j \g U)$ is a point mass.
\end{definition}

The concept of the smallest $\sigma$-algebra defines the sense in
which the variable $U$ is ``multi-cause.''  If $U$ contains
information about a single cause then it is not the smallest
separating $\sigma$-algebra.  (\Cref{app:algebra} shows why.)

The following assumption was suggested by Imai and Jiang at JSM 2019.
Here $X$ is a set of observed covariates and $Y(\mba)$ are potential
outcomes.

\begin{assumption}[No unobserved single-cause confounders]
  \label{assumption:single-cause-ignore}
  There exists a random variable $U$ that satisfies the following two
  requirements:
  \begin{enumerate}
  \item It is a multi-cause separator.
  \item Together with the observed covariates $X$, it satisfies weak
    unconfoundedness,
    \begin{align}
      \label{eq:factor-ignore}
      A_{1}, \ldots, A_{m}\indpt Y(\mathbf{a}) \g U, X &&
      \forall \mathbf{a}\in\mathcal{A}.
    \end{align}
  \end{enumerate}
\end{assumption}
The first part ensures that $U$ only involves multiple causes.  The
second part ensures that the variable $U$ contains all multi-cause
confounders.  (It can contain other multi-cause variables as well.)

Why is \Cref{assumption:single-cause-ignore} called ``no unobserved
single-cause confounders''?  The variable $U$ is a multi-cause
separator: it cannot capture single-cause variables; it must capture
ancestors of multiple causes; and it cannot capture descendants of
multiple causes.  For $U$ and the observed covariates $X$ to satisfy
weak unconfoundedness, the observed covariates must include all
single-cause confounders.

The next assumption is that every multi-cause separator is pinpointed
by a single function of the observed causes.  It is called a
substitute confounder, though only after the theorem below will it
deserve this name.
\begin{assumption}[The substitute
  confounder] \label{assumption:consist-subconf} All multi-cause
  separators $Z$ are pinpointed by a single deterministic function of
  the causes,
  \begin{align}
    \label{eq:consist-subconf}
    \upP(Z\g A_{1}, \ldots, A_{m}) = \delta_{f(A_{1}, \ldots, A_{m})},
  \end{align}
  where $\delta_{f(\cdot)}$ denotes a point mass at $f(\cdot)$.
\end{assumption}

\begin{theorem}[Weak unconfoundedness for the substitute confounder]
  Suppose
  \Cref{assumption:single-cause-ignore,assumption:consist-subconf}
  hold.  Consider a multi-cause separator $Z$. It satisfies weak
  unconfoundedness,
  \begin{align}
    A_{1}, \ldots, A_{m}\indpt Y(\mba) \g Z, X &&
    \forall \mathbf{a}\in\mathcal{A}.
  \end{align}
\end{theorem}

\Cref{sec:proof} proves the theorem.
\Cref{assumption:single-cause-ignore} posits the existence of a
multi-cause separator that also satisfies unconfoundedness.
\Cref{assumption:consist-subconf} implies there is only one
multi-cause separator $Z$, it is unique. (The next paragraph discusses
why.)  These two assumptions together imply that the multi-cause
separator $Z$ also satisfies weak unconfoundedness.

\Cref{assumption:consist-subconf} pinpoints the separator as a
function of the causes.  Why does this assumption imply the uniqueness
of the multi-cause separator, particularly across the probability
space expanded with all potential outcomes?  \Cref{eq:consist-subconf}
implies
\begin{align}
  \label{eq:full_conditional}
  \upP(Z \g A_1, \ldots, A_{m},
  \{Y(\mathbf{a})\}_{\mathbf{a}\in\mathcal{A}}, X) =
  \delta_{f(A_{1},
  \ldots, A_{m})}.
\end{align}
If two variables $Z_1$ and $Z_2$ both satisfy
\Cref{eq:full_conditional} then they must be equal.  The reason is
that $Z_1 = Z_2 = f(\mbA)$ in the full probability space
$\mathcal{A}\times\mathcal{X}\times\{\mathcal{Y}(\mathbf{a})\}_{a\in\mathcal{A}}$.
Thus \Cref{assumption:consist-subconf} implies the uniqueness of the
multi-cause separator.

The identification theorems of \citet{Wang:2019a} rely on the weak
unconfoundedness of the substitute confounder $Z$ and consider
adjustments as if $Z$ were explicitly observed.  The theorems rest on
further technical assumptions because a pinpointed $Z$ violates
overlap of the whole set of causes $\mbA$. (These technicalities are
distinct from the fact that $Z$ satisfies weak unconfoundedness.)
Note that if $Z$ is not pinpointed then there is uncertainty about the
separator, even with infinite data.  But with further assumptions,
point identification is still possible.  See \citet{Wang:2019a}
(Appendix C) and \citet{imaicomment}.\footnote{The theory in this
  paper is the same as in \citet{Wang:2019a,Wang:2019b} (WB), though
  it is clarified here.  For readers interested in the mapping: WB
  Definition 4 says there exists a smallest-$\sigma$-algebra variable
  that renders the causes independent (WB Eq 40), and it satisfies
  weak unconfoundedness.  For weak unconfoundedness, fix WB Eq 39 by
  adding $A_{i,-j}$ to its conditioning or, equivalently, use Imai and
  Jiang's assumption articulated in \citet{Wang:2019b}.  WB Definition
  5 says that there the substitute confounder is pinpointable through
  a probabilistic factor model; because it's probabilistic, none of
  the factors is a point mass.  Finally, the proof of Theorem 1 in
  this paper appears at the end of the proof of WB Lemma 2.}

\section{From the theory to the algorithm}

The deconfounder algorithm of \cite{Wang:2019a} operationalizes this
theory.  \textit{If} the investigator finds a deterministic function
of the causes that renders them conditionally independent
\textit{then} the output of that function can be used as a substitute
confounder in a downstream causal inference.
\Cref{assumption:single-cause-ignore} is that the observed covariates
$X$ and multi-cause separator $U$ provide weak unconfoundedness: there
are no unobserved single-cause confounders.
\Cref{assumption:consist-subconf} is that there is a single $f(\mbA)$
that provides the substitute confounder.

The algorithm uses a probabilistic factor model and posterior
predictive check to find $f(\mbA)$.  Suppose a factor model describes
well the distribution of the causes. Then its local latent variable
renders the causes conditionally independent.  When the number of
causes is large and the local latent variable is low-dimensional, this
inference approaches a deterministic function, satisfying
\Cref{eq:consist-subconf}.  The deconfounder infers the local latent
variables and calls them substitute confounders.

Why are the inferred confounders multi-cause separators?  Why do they
form the smallest $\sigma$-algebra that renders the causes
conditionally independent?  The reason has two parts. (1) The factor
model implies that its latent variable renders the causes
conditionally independent. (2) The $\sigma$-algebra of a pinpointed
separator cannot pick up single-cause variables; \citet{Wang:2019b}
provides a proof.

\Cref{assumption:consist-subconf} requires a single deterministic
function that provides the separator; the algorithm uses a factor
model to find it.  \citet{Kruskal:1989} and \citet{Allman:2009}
provide conditions that guarantee the uniqueness of the factor model
that captures the distribution of the causes. \citet{Bai:2012} and
\citet{Chen:2019} study conditions under which the latent variables of
factor models are identifiable.  With many causes and a
low-dimensional factor model, inference of its local variable
approaches a deterministic function.

Finally, the deconfounder uses posterior predictive
checks~\citep{Guttman:1967,Rubin:1984,Gelman:1996} to assess the
fidelity of the distribution of causes that is provided by the factor
model.  Specifically, the check evaluates the predictive distribution
on sets of held-out causes. This strategy uses ideas from
\citet{Bayarri:2000,Robins:2000b,Ranganath:2019} to provide a
better-calibrated check.

\section{The proposed counterexamples violate the assumptions}

\citet{Ogburn:2020} propose counterexamples to the theory of
\citet{Wang:2019a}.  Using the theory, as outlined above, the proposed
counterexamples do not satisfy the required assumptions.

\parhead{Example 1.} There are two independent causes $A_1$ and $A_2$
and a substitute confounder $Z\sim \textrm{Bernoulli}(0.5)$ that is
independent of all other variables $(A_1, A_2, Y(\mathbf{a}))$. Here
the substitute confounder $Z$ does not satisfy
\Cref{assumption:consist-subconf}; its conditional distribution
$\upP(Z\g A_1, A_2)=\textrm{Bernoulli}(0.5)$ is not a point mass.

\parhead{Example 2.} There are two causes $A_1$ and $A_2$, and another
variable $U$. Assume that $A_j \indpt Y(\mba) \g U, j=1, 2$ and $U$ is
the smallest $\sigma$-algebra that renders $A_1$ and $A_2$
conditionally independent. Set the substitute confounder $Z=U$.  Again
$\upP(Z \g A_1, A_2)$ is not a point mass, violating
\Cref{assumption:consist-subconf}.  Further, $U$ does not satisfy weak
unconfoundedness because $(A_1, A_2) \not \indpt Y(\mba) \g U$, violating
\Cref{assumption:single-cause-ignore}.\footnote{This example does
  satisfy the ``no unobserved single-cause confounder'' assumption as
  stated in Definition 4 of \citet{Wang:2019a}. But it violates the
  refined assumption in \Cref{assumption:single-cause-ignore}, due to
  Imai and Jiang at JSM and also published in in \citet{Wang:2019b}.}

\parhead{Example 3.} A variant of the second example involves a third
cause $A_3$ and sets the substitute confounder $Z=A_3$. This example
violates \Cref{assumption:consist-subconf}.  A pinpointed substitute
cannot be a function of only one cause~\citep{Wang:2019b}.

\parhead{The importance of conditional independence.}
\citet{Ogburn:2020} claim that the conditional independence
requirement of factor models does not ``drive the success'' of
deconfounder-like methods.  But the conditional independence
requirement, along with pinpointability, plays an important role in
confirming the assumptions required by the algorithm.
\begin{enumerate}[leftmargin=*]
\item Requiring conditional independence outlines the class of
  confounders that the deconfounder targets; they must be multi-cause
  confounders.  This requirement is why, with the assumption of no
  unobserved single-cause confounder, the deconfounder handles all
  confounders.

\item Requiring conditional independence prevents the substitute
  confounder from capturing multi-cause colliders or mediators;
  capturing such variables violates the conditional independence
  requirement.

\item As for single-cause post-treatment variables, \citet{Wang:2019b}
  shows that substitute confounders that satisfy
  \Cref{assumption:consist-subconf} cannot capture any single-cause
  variables.  Thus, along with point \#2 above, the substitute
  confounder does not capture any post-treatment variables.
\end{enumerate}

\section{Discussion}

To reiterate \citet{Wang:2019b}, the deconfounder is not a turnkey
solution to causal inference. It does not relieve the researcher from
trying to measure confounders. As for all causal inference with
observational data, it comes with uncheckable assumptions.  In
particular, \Cref{assumption:single-cause-ignore} is that there are no
unobserved single-cause confounders.

This paper clarifies the theoretical foundations of
\citet{Wang:2019a}.  The refinements of Imai and Jiang make clearer
the assumptions required for identification, and they help simplify
the proof that the substitute confounder satisfies weak
unconfoundedness.

Both \citet{ogburncomment} and \citet{Ogburn:2020} question the
correctness of the theory behind the deconfounder.  The objections are
incorrect, discussed above and in \citet{Wang:2019b}.  But more
broadly, \citeauthor{ogburncomment}'s resistance to the idea stems
from an important misunderstanding.  Both commentaries reiterate the
fact that no information about unobserved confounders can be inferred
from observational data. \citet{Wang:2019a} does not challenge this
fact.  Rather, the theory finds confounders that are
\textit{effectively observed}, even if not explicitly so, and embedded
in the multiplicity of the causes.  The deconfounder extracts this
information for causal inference.

\section*{Acknowledgements}

We thank Don Green, Kosuke Imai, Aaron Schein, and Brandon Stewart for
useful discussions.

\appendix

\section{Smallest $\sigma$-algebra}
\label{app:algebra}

The restriction of the smallest $\sigma$-algebra ensures that $U$ can
only pick up multiple-cause variables. The proof is by contradiction:
if $U$ involves both multi-cause and single-cause variables then $U$
cannot be the smallest $\sigma$-algebra.

Formally, suppose the variable $U$ contains a single cause component
and a multi-cause component, $U = (U^\rms, U^\rmm)$.  Without loss of
generality, suppose the single-cause component only depends on the
first cause $A_1$.  Then \Cref{eq:factor-condindep} implies
\begin{align}
  \upP(A_{1}, \ldots, A_{m}\g U)
  &= \prod_{j=1}^m \upP(A_{j}\g U)\\
  &= \upP(A_{1}\g U^\rms, U^\rmm)\cdot \prod_{j=2}^m \upP(A_{j}\g
    U^\rmm).
\end{align}
This implies that
\begin{align}
  \upP(A_{1}, \ldots, A_{m}\g U^\rmm)
  &= \int \upP(A_{1}, \ldots, A_{m}\g U) \cdot \upP(U^\rms \g U^\rmm) \dif U^\rms\\
  & = \prod_{j=2}^m \upP(A_{j}\g U^\rmm) \cdot \int \upP(A_{1}\g
    U^\rmm, U^\rms)\cdot \upP(U^\rms\g U^\rmm)\dif U^\rms\\
  & = \upP(A_{1}\g U^\rmm)\cdot
    \prod_{j=2}^m \upP(A_{j}\g U^\rmm),
\end{align}
which means that $U = (U^\rms, U^\rmm)$ is not the smallest
$\sigma$-algebra that renders the causes independent.

\section{Proof of Theorem 1}
\label{sec:proof}

\Cref{eq:consist-subconf} implies the following conditional
distribution of the substitute confounder given all of the other
variables (including the potential outcomes),
\begin{align}
\label{eq:full-pinpoint}
\upP(Z \g A_{1}, \ldots, A_{m},
\{Y(\mathbf{a})\}_{\mathbf{a}\in\mathcal{A}}, X) =
\delta_{f(A_{1},
\ldots, A_{m})}.
\end{align}
\Cref{assumption:consist-subconf} only concerns the probability space
of the observed causes $\mathcal{A}$.  But \Cref{eq:full-pinpoint}
holds because $\upP(Z \g A_{1}, \ldots, A_{m})$ is a point mass
$\delta_{f(A_{1}, \ldots, A_{m})}$, which satisfies
$\delta_{f(A_{1}, \ldots, A_{m})} \indpt
\{Y_i(\mathbf{a})\}_{\mathbf{a}\in\mathcal{A}}, X$.

\Cref{eq:full-pinpoint} implies that the substitute confounder $Z$ is
unique.  The reason is that if two variables $Z_1$ and $Z_2$ satisfy
\Cref{eq:full-pinpoint} then they must be equal in the whole
probability space, $Z_1 = Z_2 = f(A_{1}, \ldots, A_{m})$ on
$\mathcal{A}\times\mathcal{X}\times\{\mathcal{Y}(\mathbf{a})\}_{a\in\mathcal{A}}$.

The uniqueness of $Z$, together with
\Cref{assumption:single-cause-ignore}, implies that the substitute
confounder satisfies weak unconfoundedness,
\begin{align}
  A_{1}, \ldots, A_{m} \indpt Y_i(\mathbf{a}) \g Z, X
  &&
     \forall \mba \in \mathcal{A}.
\end{align}
Why?  \Cref{assumption:single-cause-ignore} asserts the
\textit{existence} of a random variable that is (1) the smallest
$\sigma$-algebra that renders the causes conditionally independent and
(2) satisfies weak unconfoundedness. The argument above establishes
the \textit{uniqueness} of the random variable that satisfies
(1). Thus the variable satisfying (1) must also satisfy (2).  This
unique random variable is the substitute confounder $Z$.

Note that this proof simplifies the proof of Lemma 2 in
\citet{Wang:2019a}.

\bibliographystyle{apalike}
\bibliography{bib.bib}

\end{document}